# Policy Fragmentation or Institutional Alignment?
# Institutional Governance of AI in Universities and Business Schools

**Lydia Manikonda**
Lally School of Management
Rensselaer Polytechnic Institute
Troy, NY 12180
manikl@rpi.edu

**Dominique Outlaw**
Department of Finance
Frank G. Zarb School of Business
148 Hofstra University,
Hempstead, NY 11549
dominique.outlaw@hofstra.edu

**Abstract**

Artificial intelligence (AI) is rapidly transforming high-skilled domains, requiring higher education institutions (HEI) to balance the teaching of foundational principles with the integration of emerging tools to ensure workforce readiness. While HEI are increasingly adopting AI, many continue to grapple with how it should be incorporated into curricula and governed through policy, especially when such policies are set at different levels of an institution. This research analyzes AI policies across HEI from 34 states in the United States to investigate what these policies entail and how policies set across institutions as well as within different levels at an institution differ. Using natural language processing (NLP) to analyze institutional AI policies, we find a clear divergence: university-level policies emphasize data security and risk mitigation whereas school-level policies, when present, focus on pedagogical applications and tool usage. When focusing on business school specific policies, relatively few business schools maintain AI policies distinct from university frameworks, creating misalignment with discipline-specific learning objectives. This gap poses challenges particularly for faculty and students as well as for accreditation purposes. Our insights suggest that guidelines should be aligned with broader institutional policies while addressing discipline-specific learning objectives and evolving workforce demands.

## Introduction

Emerging technologies have consistently reshaped the academic landscape, often prompting debate on policies regarding their appropriate role in teaching and assessment. In higher education, these debates are not new. Decades ago, instructors questioned whether calculators should be permitted in the classroom, particularly during examinations. Over time, however, calculators became normalized with textbooks incorporating instruction on their use and professional certification bodies establishing explicit guidelines governing their application. A similar trajectory occurred with spreadsheet software that is now widely regarded as a baseline expectation for students and a prerequisite for employment in the industry. Today, artificial intelligence (AI) represents the next, and arguably most transformative, phase in this technological.

The rapid adoption of AI tools across the financial services sector, including applications in forecasting, risk management, algorithmic trading, and financial analysis has fundamentally altered employer expectations (Brynjolfsson and McAfee 2017; Agrawal, Gans, and Goldfarb 2019). Increasingly, firms expect graduates not only to possess strong foundational knowledge in finance but also to demonstrate the ability to effectively leverage AI tools to augment decision-making and improve productivity. In this context, AI literacy is quickly becoming a non-negotiable competency for entry-level finance professionals (Dougherty and Smith 2025; Tarnowski 2025).

Despite this clear shift in industry expectations, higher education institutions (HEI) have struggled to determine the appropriate role of AI within teaching and assessment. On one hand, AI tools offer opportunities to enhance learning, support individualized instruction, and better align educational outcomes with industry needs (Liu, Zhou, and Matsudaira, 2025; Córdova et al. 2024). On the other hand, concerns persist regarding academic integrity, overreliance on automation, and the potential erosion of critical thinking and problem-solving skills (Kasneci et al. 2023). This tension has created significant uncertainty among faculty and administrators, particularly in disciplines such as finance, in which both conceptual rigor and technical proficiency are essential.

This article seeks to address this gap by examining the current landscape of AI-use policies in HEI with a particular focus on analyzing university-wide AI policies across institutions and then comparing business school's AI policy to the university-wide policy within the same institution. First, we provide a systematic overview of existing institutional policies, highlighting common approaches, areas of convergence, and points of divergence. As many institutions remain in the early stages of policy development, such an overview offers timely insights and practical guidance for educators and administrators navigating similar challenges. While prior work has begun to explore the integration of generative AI into higher education, particularly through frameworks such as constructive alignment (Zhou et al. 2026), comprehensive analyses of policy frameworks, highlighting similarities and differences in policy focus across institutions remain limited. Second, we examine a sample of R1 universities to identify discrepancies between university-level policy and those implemented at the business school-level within the same institution. In many cases, institutional policies provide broad guidance while individual schools or departments develop more specific rules tailored to disciplinary needs. This layered governance structure can result in inconsistencies that create confusion for faculty and students. Understanding these differences is especially important in finance education, in which the integration of technology is not only pedagogically relevant but also professionally necessary.

This article proposes a set of guiding principles and policy recommendations for AI use in higher education and ways  to bridge the gap between policies set within within a given school policy compared to its university-wide policy. These recommendations aim to strike a balance between fostering innovation and maintaining academic integrity while also ensuring that students develop the critical thinking skills necessary to evaluate and responsibly use AI-generated outputs. The goal is not to prescribe a one-size-fits-all solution,but rather to offer a framework that institutions can adapt to their specific contexts. The importance of this issue extends beyond individual classrooms. HEI are increasingly expected to produce workforce-ready graduates who can thrive in technology-driven environments and especially with strong evidence suggesting that education has a direct influence with industry growth, e.g., finance education (AlSuwaidi and Mertzanis 2024). When focusing within an institution, business schools must meet the standards of accrediting bodies such as the Association to Advance Collegiate Schools of Business (AACSB), which emphasize the integration of emerging technologies and the development of relevant, future-oriented curricula (AACSB 2020). As such, schools must be able to demonstrate not only that AI is incorporated into their programs but also that its use is guided by clear, coherent, and pedagogically sound policies coherent with the policies set at a university-level.

**Policy Debate and Existing Work**

There is a central policy debate about AI in terms of how to capture economic and scientific benefits while managing real-world risks to society (Suter et al. 2025). Within higher education, concerns have largely moved away from existential AI risks and toward its implications for teaching, learning, assessment, and institutional governance. (Delaney and Odle 2025; Ye 2024). Universities and business schools are actively grappling with a central question related to policies: *how should AI use be governed in academic settings*? Further, how can domain-specific programs such as finance programs design AI-use policies that align with broader institutional governance frameworks while preserving the rigor, analytical depth, and ethical standards traditionally associated with these disciplines and meeting industry standards? The challenge is not merely whether to allow AI but how to integrate it in a manner that supports learning objectives. Thus, we investigate this research question "*how do HEI govern AI, and to what extent do individual academic units develop policies that differ from university-wide guidelines?*" To our knowledge, this study is the first to systematically collect, quantitatively analyze, and characterize AI-use policies across universities via sentiments and themes of focus and further examine whether decentralized academic units such as business schools develop policies that differ from the institution's central policy.

**Data**

We collect AI policies from 149 Carnegie Classification of Institutions of Higher Education classified as R1 or R2 in the United States. We use an automated Python-based crawler to collect the content of the AI policies that are in the form of natural language text. We then manually verify each AI policy to ensure the crawler did not extract irrelevant information. We remove any potentially erroneous entries to obtain a dataset of 130 policies as the university-focused corpus. These policies cover universities from 34 different states, suggesting that it is representative of the HEI around the country. Of the 130 universities, we found only eight business schools that have a school-specific policy at the time of crawling this dataset, and we crawl AI policies for these business schools. These business schools are Haas School of Business at University of California- Berkeley, Columbia Business School, Tuck School of Business at Dartmouth College, Harvard Business School, MIT Sloan School of Management, Kellogg School of Management at Northwestern University, Ross School of Business at University of Michigan, and the Wharton School at University of Pennsylvania.

## Empirical Strategy

To objectively analyze and compare the AI policies, we use sophisticated modeling techniques to construct three different measures around the sentiment of the policy, the clarity of the policy, and the similarity of policies, which we describe in this section.

*Sentiment Polarity*: Sentiment polarity gives an understanding of how a particular corpus or a piece of text expresses emotional tone that could be positive, negative or neutral. The range of the polarity values typically are on a scale of -1 to 1 where, -1 means negative and 1 meaning positive with 0 being neutral in sentiment. Computing sentiment polarity values (via Spacy Python Library) provides an understanding of the latent emotional tone expressed via natural language content. We leverage sentiment polarity values to identify whether AI policies are enthusiastic or pessimistic.

*Topic Modeling*: We then extract topics using topic modeling via neural embeddings via BERTopic approach (Grootendorst 2022) that captures both semantics and syntax while identifying the topics focused by the extracted policies. Topic modeling extracts the word distributions across each topic, and we synthesize the topics using the keywords present in the topic distribution. We leverage Python-based API to perform the topic extraction. We also investigate the language used in the policies in terms of whether the language is restrictive or unrestrictive by focusing on the verbs included in the policy text. To do this, we first extract part of speech tags, specifically verbs and auxiliary verbs for university and business school policies.

*Cosine Similarity via Sentence Embeddings*: We then perform vectorization of policies using sentence embeddings approach via SBERT (Reimers and Gurevych 2019) and compute the cosine similarity to measure the similarity between policies. If two policies are semantically and syntactically similar, the cosine similarity will be 1; otherwise, 0. Thus, we have a more meaningful way to quantitatively evaluate natural language text in policies as these approaches consider syntax, semantics, and context.

**Current AI Policies in Higher Education: A Systematic Evaluation**

*Across-Institution Comparison: University-Wide policies*:

We examine sentiment polarity and subjectivity to characterize the overall tone of AI policies and their stance toward AI adoption.

[*Insert Figure 1(a) Here*]

Polarity values in Figure 1(a) highlight that most universities are positive in sentiment; however, they are closer to 0 than 1, suggesting that the majority are moderately positive. Only two institutions exhibit a negative tone and both are near neutral, indicating no strong opposition to AI. The distribution of these polarity values suggests that universities are not taking a concrete stance in terms of providing specific guidelines and are carefully treading this rapidly changing landscape of AI. For subjectivity scores, 0 means fully objective and 1 means fully subjective. The distribution in Figure 1(b) suggests that 80% of the policies are mostly neutral (with neither being highly subjective nor highly objective) with a subjectivity score of >0.4 (*mean*=0.452). This is logical as these AI policies curated for specific institutions are presented as suggested guidelines.

[*Insert Figure 1(b) Here*]

Next, we assess policy clarity by comparing the use of strong directive language (e.g., ‘must,’ ‘prohibited’) with weaker, suggestive language (e.g., ‘may,’ ‘can’).Strength is measured by the difference of the total number of stronger words and weaker words, suggesting that if the strength value is negative, weaker words are dominating in frequency in a given policy. Figure 2 shows that 95% of policies have a negative strength score suggesting the usage of weaker and less assertive language, which corroborates with the insights we identified from polarity values. Most universities take a careful stance, using non-emphatic language to implement AI.

[*Insert Figure 2 Here*]

We then extract the focused themes expressed in the policies using the neural embedding-based BERTopic topic modeling approach to have a high-level understanding of what these policies entail and what key areas are they focusing on. The key themes are included as part of the Appendix A and can be summarized into three main categories: 1) usage and education, 2) technology and tools, and 3) governance and potential risks. However, as we saw from the clarity strength, use of language in these policies is overwhelmingly cautious and intended as guidance but not a mandate. This reticence to be specific may explain why several of the policies are shared with the university community as "guidelines" rather than "policy."

*Within-Institution Comparison: Business school-focused policies*

To examine policy fragmentation within institutions, we analyze the AI policies of business schools and compare them to the corresponding university's policy. To do this, we follow a similar process and convert each policy into neural embeddings (Reimers and Gurevych 2019) and compute the cosine similarity value. Table B1 in Appendix B presents the summary statistics of similarity values. The values of cosine similarity are in the range of [0,1]. Except for UC-Berkeley and Northwestern, the business school policies differ from the university.

To further investigate, Tables B2 and B3 in Appendix B present the school and university-wide differences in policies. Business schools focus on tools and agreement, specifications on course expectations for faculty, guidelines for students, AI risks and compliance, and AI literacy and skill development.

The corresponding university-wide policies focus on similar areas but additionally on guidelines for procurement, confidential research protection, AI detection in courses, common guidelines for both faculty and students on academic integrity and tutoring and academic support with AI. Of note, most of the school-level policies focus on different tools, accessing those tools, and model outputs whereas at the university level, the main focus is data security and procurement. Additionally, school-level policies focus on pedagogy while university-level policies primarily focus on risk management (Figure 3).

[*Insert Figure 3 Here*]

## Alignment with Accreditation Standards for Business Education

The topics extracted from business school-specific and university-wide policies reveal differing priorities, highlighting potential tensions between centralized institutional governance and local instructional objectives. Because AI adoption is still in a transition state, most institutions are not yet meeting accreditation expectations of unified integration of curriculum, policy, assessment, and infrastructure. Today, most universities take a moderately positive stance toward AI adoption (moderately positive polarity). They are primarily concerned with the safe and responsible use of AI whereas business schools often focus on instructional value. However, policies that support student competency development in emerging technologies must move beyond questions of safety and adoption to address how AI is integrated into teaching and learning. ***Guidelines should be aligned with broader institutional policies while addressing discipline-specific learning objectives and evolving workforce demands.***

**Proposed Policy Framework**

Policies are most effective when they are designed in concert with learning objectives and pedagogical practices. As we learn from the results in the previous section, instead of focusing on data security and procurement standalone, integrating procurement needs with teaching needs is a must. Concurrently, it is important to standardize disclosure across all courses in the department, school, and universities with a continuous feedback loop where teaching and pedagogy informs the policy and vice versa. Regular policy review is essential as technological capabilities, and workforce demands continue to evolve.

We recommend starting with the University-wide risk management policy. The purpose of this foundational layer is to ensure the policy maintains consistency, legal compliance, and academic integrity across all disciplines; therefore, the university should set up some guardrails with a specific focus on ensuring students are workforce-ready. To this end, policy must consider maintaining transparency as well as redefining aspects around academic integrity such as providing clear definitions around plagiarism. Most importantly, they should provide clear guidance on exercising caution when uploading data to AI platforms especially, setting up strict rules prohibiting the upload of sensitive information. When it comes to the level of each department at an institution, by honoring the university-wide guidelines, there should be unit-specific contextual policies that help foster student learning and deliver high-quality education that mitigates confusion. By drawing on the insights identified in this study, we suggest there should be stronger collaborative mechanisms between academic units, such as inter-department policy committees comprising faculty at different levels across units as well as students across disciplines while continually revising these policies to address policy fragmentation.

## Limitations and Future Research

The investigation presented in this paper focuses on a set of only 130 institutions across the United States. Even though they represent 34 different states highlighting geographic diversity, this sample size may limit generalizability. The insights suggest potential for broader application and further research should examine including diverse disciplines and student populations to establish more comprehensive understanding of how AI policies are being proposed and implemented. Further, several institutions use the term “guidelines” rather than “policy,” leading us to question whether these statements are mandatory or optional. Future research could explicitly model this distinction through separate analytical samples.

## Conclusion

This paper contributes to the emerging literature on AI in higher education in several important ways by analyzing AI-specific policies/guidelines followed by HEI across the US. First, it provides a systematic examination of AI-use policies across institutions that highlights the lack of stronger and more assertive language to provide clear guidelines on the use of AI in higher education. Second, it adds to the very limited literature on AI in business education and bridges a gap between institutional governance and disciplinary pedagogy by comparing university-level AI policies with those developed within business schools, highlighting areas of alignment and inconsistency that have direct implications for curriculum design and student learning. Third, the paper advances the conversation beyond descriptive analysis by proposing a set of guiding principles for AI policy development that are grounded in both pedagogical theory and industry expectations. Finally, it situates AI policy within the broader trajectory of technological adoption, offering a conceptual lens to understand and respond to ongoing and future disruptions.

Existing policies are still ambiguous in terms of when and how to use these tools appropriately and efficiently. These gaps can be filled by bringing both faculty and students into the loop while designing AI policies and continuously iterating AI policies with a unified focus across university entities. Such an approach can contribute to improved student success outcomes and more effective delivery of education, particularly for populations facing barriers to academic achievement. In fact, these ambiguous and risk management-oriented institutional policies can benefit HEI to focus on more than traditional risk and include technological, legal, and governance risks, thereby encouraging pedagogical focus towards compliance-aware decision-making. We hope these insights highlight the critical role of effective AI policy in aligning educational practice with evolving workforce needs, that will strengthen both student outcomes and industry innovation.

**Declaration of generative AI use**

*The authors report that generative AI was not used in their research or preparation of this manuscript.*

**References**

AACSB. (2020). *2020 Guiding Principles and Standards for Business Accreditation*. Association to Advance Collegiate Schools of Business.

Agrawal, A., Gans, J., & Goldfarb, A. (2022). *Prediction machines, updated and expanded: The simple economics of artificial intelligence*. Harvard Business Press.

AlSuwaidi, Rashed A., and Christos Mertzanis. "Financial Literacy and FinTech Market Growth Around the World." *International Review of Financial Analysis* 95 (2024): 103481.

Brynjolfsson, Erik, and Andrew McAfee. *The Business of Artificial Intelligence*. Boston: Harvard Business Review Press, 2017.

Córdova, Paola, Alejandro Grájeda, Juan P. Córdova, Antonio Vargas-Sánchez, Jorge Burgos, and Andrea Sanjinés. "Leveraging AI Tools in Finance Education: Exploring Student Perceptions, Emotional Reactions and Educator Experiences." *Cogent Education* 11, no. 1 (2024): 2431885.

Delaney, Jennifer A., and Taylor K. Odle. "State-level common application policies and college enrollment." *Education Finance and Policy* 20, no. 1 (2025): 164-177.

Dougherty, Shaun M., and Mary M. Smith. "At What Cost? Is Technical Education Worth the Investment?." *Education Finance and Policy* 20, no. 1 (2025): 85-109.

Erkkilä, Tero. "Global Indicators and AI Policy: Metrics, Policy Scripts, and Narratives." *Review of Policy Research* 40, no. 5 (2023): 811–39.

Grootendorst, Maarten. "BERTopic: Neural Topic Modeling with a Class-Based TF-IDF Procedure." *arXiv*. March 11, 2022. https://arxiv.org/abs/2203.05794.

Kasneci, Enkelejda, Kathrin Sessler, Stefan Küchemann, Maria Bannert, Daryna Dementieva, Frank Fischer, and Götz Kasneci. "ChatGPT for Good? On Opportunities and Challenges of Large Language Models for Education." *Learning and Individual Differences* 103 (2023): 102274.

Liu, Vivian Yuen Ting, Rachel Yang Zhou, and Jordan Matsudaira. "Six Years Later: Examining the Academic and Employment Outcomes of the Original and Reinstated Summer Pell Grant." *Education Finance and Policy* 20, no. 1 (2025): 33-55.

Reimers, Nils, and Iryna Gurevych. "Sentence-BERT: Sentence Embeddings Using Siamese BERT-Networks." In *Proceedings of the 2019 Conference on Empirical Methods in Natural Language Processing and the 9th International Joint Conference on Natural Language Processing (EMNLP-IJCNLP)*, 3982–92. Hong Kong: Association for Computational Linguistics, 2019.

Suter, Vera, Chen Ma, Gregor Pöhlmann, and Miriam Meckel. "When Politicians Talk AI: Issue-Frames in Parliamentary Debates Before and After ChatGPT." *Policy & Internet* 17, no. 3 (2025): e70010.

Tarnowski, Randy. Workforce Outlook: Class of 2026 in the AI Economy. Handshake, August 1, 2025. https://joinhandshake.com/research/economic-research/class-of-2026-ai-outlook/

Ye, Xiaoyang. "Improving college choice in centralized admissions: Experimental evidence on the importance of precise predictions." *Education Finance and Policy* 19, no. 2 (2024): 308-340.

Zhou, Xinyue, Qing Chai, Bhargav Chilukuri, and Jennifer Quach. "From Experimentation to Integration: Embedding GenAI in Business Higher Education Through the Lens of Constructive Alignment." *Journal of University Teaching and Learning Practice* (2026). https://doi.org/10.53761/pc04tp05.

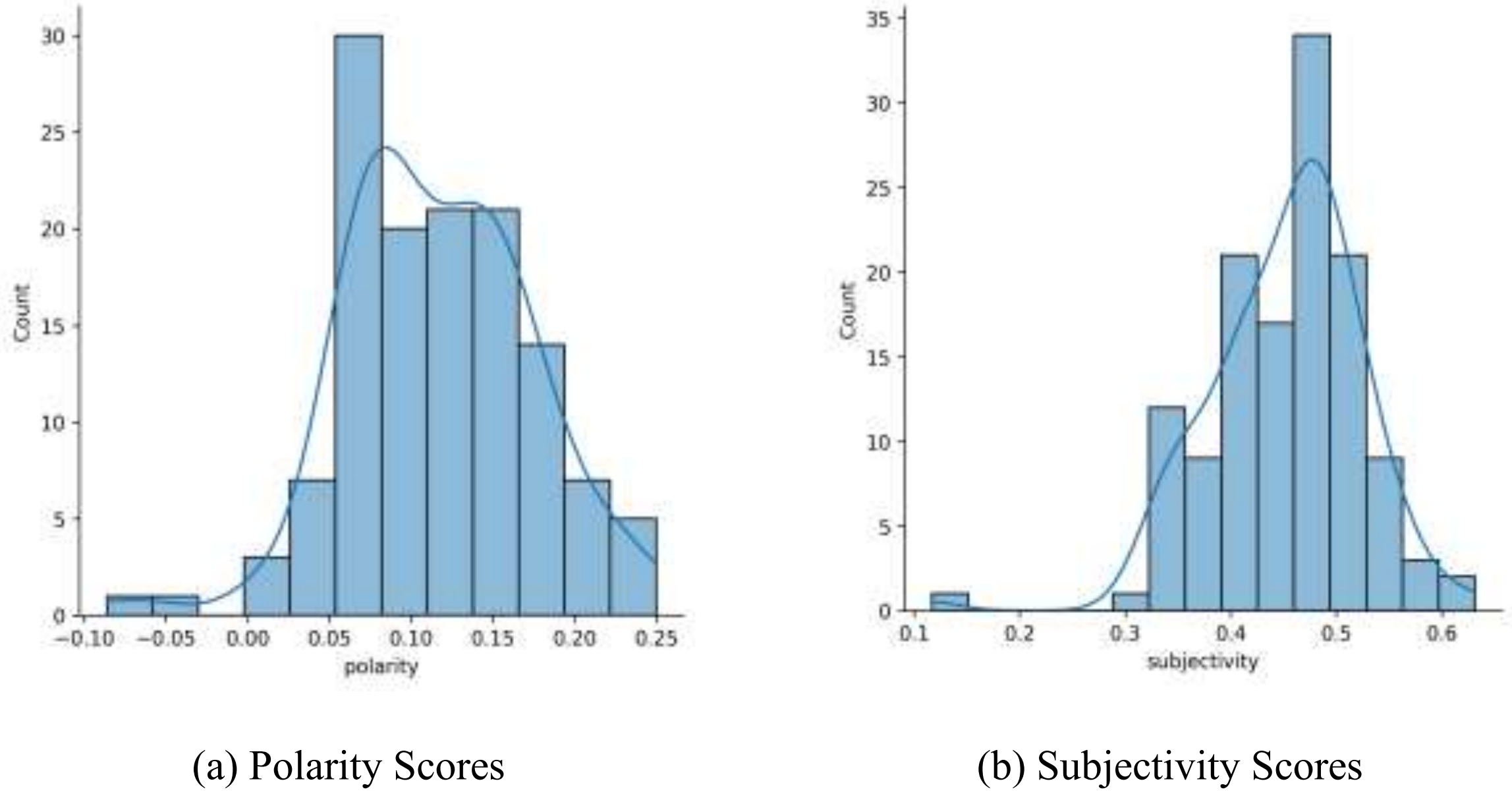

(a) Polarity Scores

(b) Subjectivity Scores

**Figure 1**: **(a)** Distribution of polarity scores (*mean*=0.11, *min*=-0.08, and *max*=0.25) extracted from the AI policy text. Sentiment polarity gives an understanding about how a particular corpus or a piece of text expresses emotional tone that could be positive, negative or neutral. The range of the polarity values typically are on a scale of -1 to 1 where, -1 means negative and 1 meaning positive with 0 being neutral in sentiment. **(b)** Distribution of subjectivity polarity scores (*mean*=0.45, *min*=0.12, and *max*=0.63) extracted from the AI policy text

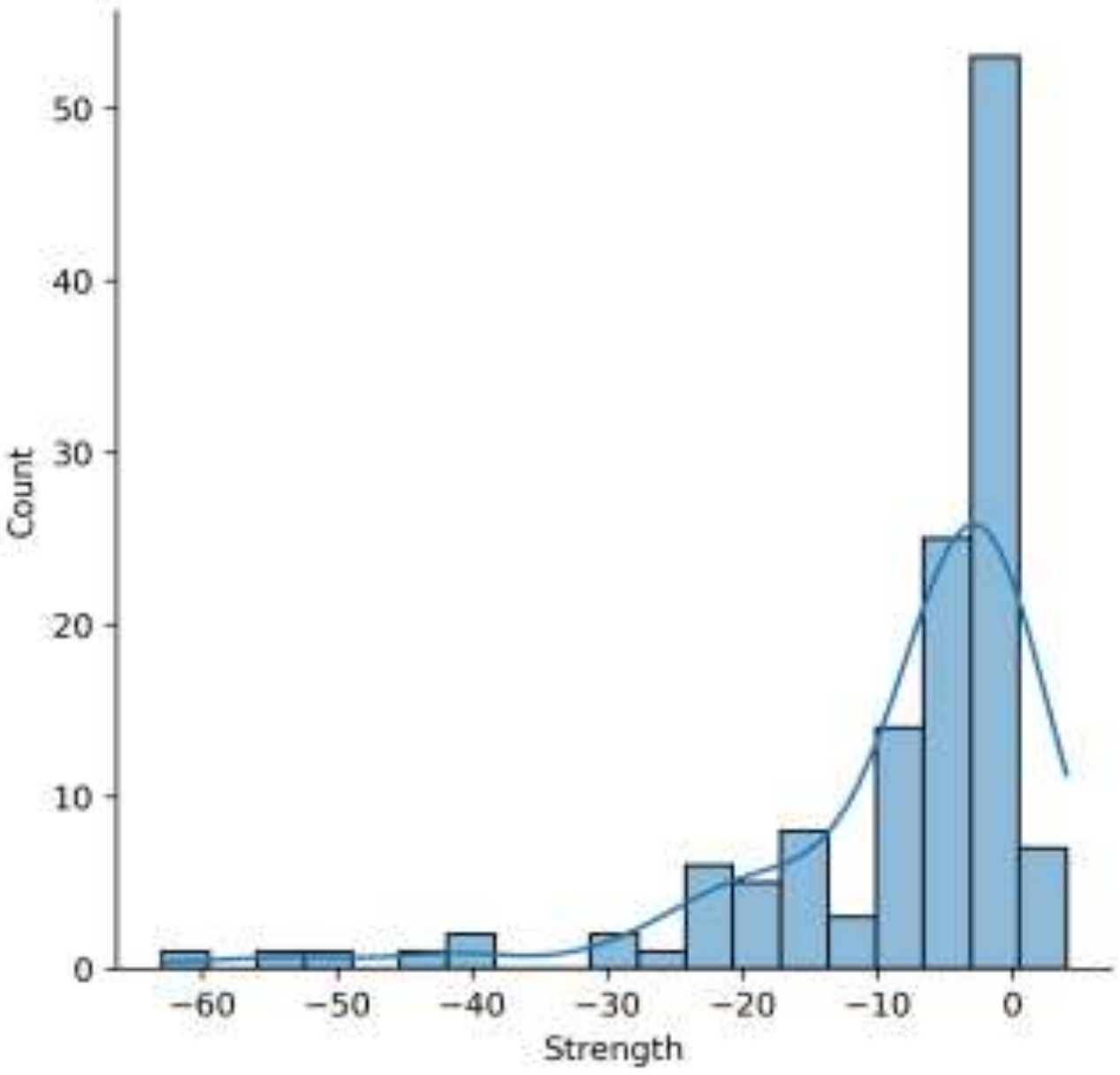

**Figure 2**: Distribution of clarity strength values (*mean*=-8.29, *min*=-63, and *max*=4) extracted from the AI policy texts. Clarity strength suggests how a policy is using stronger vocabulary such as "must", "prohibited", "should", etc., in comparison with weaker vocabulary such as "may", "suggest", "can", etc. Strength is measured by the difference of total number of stronger words and weaker words suggesting that if the strength value is negative, weaker words are dominating in frequency in a given policy.

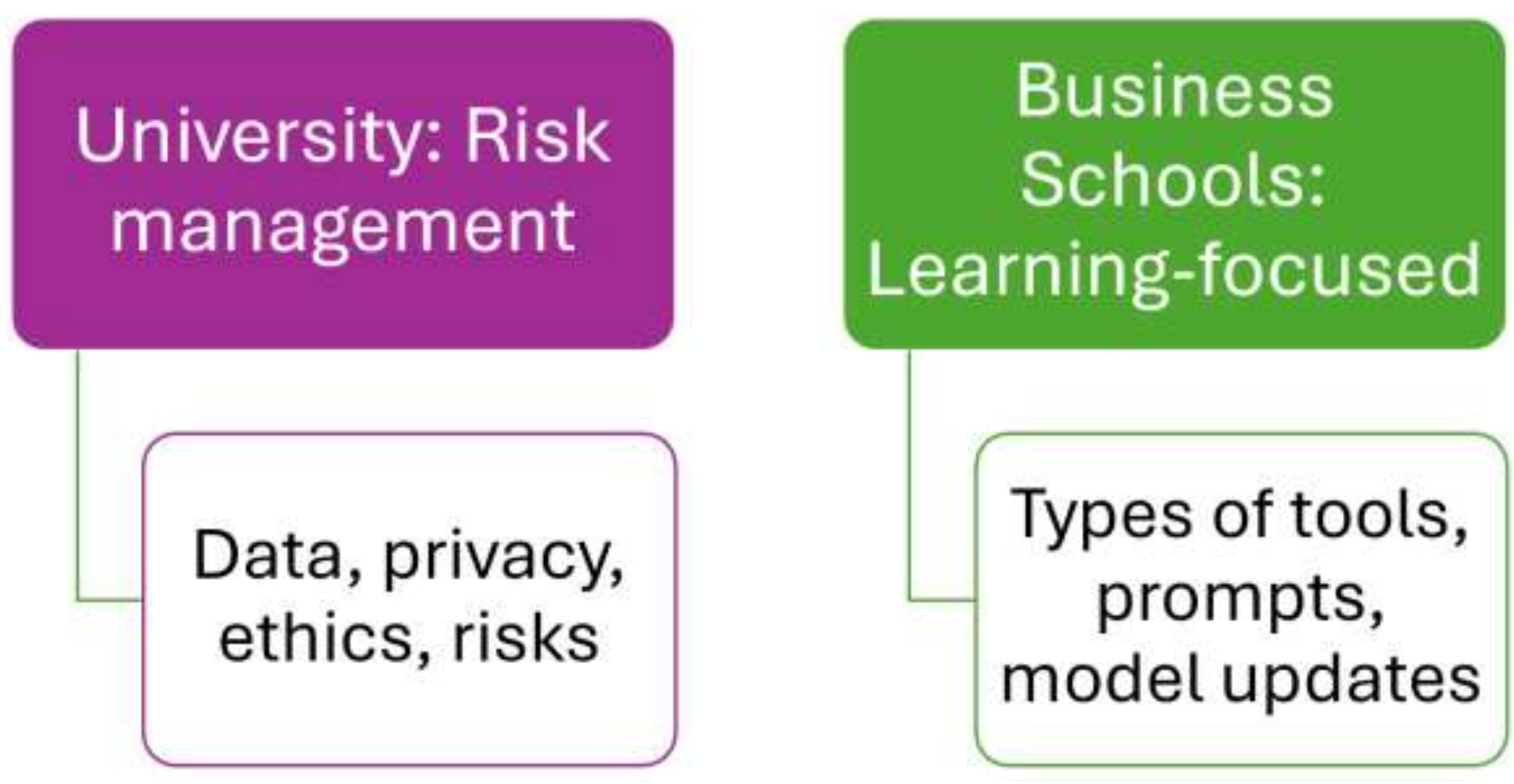

**Figure 3:** Differences between business school-specific policy vs university-wide

## Appendix A

The key themes extracted using topic modeling are (in no particular order):

1. GenAI use in courses and academic integrity
2. AI tools and platforms
3. Data privacy, security, and compliance
4. Institutional policies and governance
5. Responsible AI, risk, and ethics
6. Research, IRB, and human subjects
7. AI capabilities and concepts
8. AI in teaching and learning innovation
9. Procurement, vendors, and approved tools
10. Cybersecurity and technical infrastructure

**Appendix B**

**Table B1**: Summary statistics of the pairwise cosine similarity computed between each business school policy and their corresponding university policy. We perform vectorization of policies using sentence embeddings approach via SBERT (Reimers and Gurevych 2019) and then compute the cosine similarity to measure the similarity between policies. If two policies are semantically and syntactically similar, the cosine similarity will be 1; otherwise, 0.

| **Count** | **Mean** | **Std dev.** | **Min** | **Max** |
|---|---|---|---|---|
| 8 | 0.767 | 0.167 | 0.59 | 0.99 |

**Table B2**: Topics extracted from the text documents of AI policies at business school-level. Topic modeling extracts the word distributions across each topic, and we synthesize the name of each topic using the keywords present in the topic distribution. This particular topic modeling combines the strengths of transformers and c-TF-IDF approached to create dense clusters that allows for identifying interpretable topics while keeping important words in the topic descriptions. We leverage Python-based API to perform the topic extraction automatically.

| ID | Theme | Vocabulary | % of policies |
|---|---|---|---|
| 1 | **AI tool access and vendors** | tool, use, access, tools, external, link, pdf, file, vendor | 11.11% |
| 2 | **GenAI model outputs** | models, genai, based, output, true, using, activity, might, information, english | 11.11% |
| 3 | **Generative AI tools overview** | data, generative, learn, adobe, microsoft, tools, overview, creative, copilot, chatgpt | 11.11% |
| 4 | **Student academic use policy** | tools, academic, students, use, ai, must, seek, appropriately, content, ensure | 8.33% |
| 5 | **Faculty and course expectations** | students, generative, faculty, course, using, academy, cbs, policy, permission, expectations | 8.33% |
| 6 | **Data sharing & confidentiality** | data, level, publicly, share, information, sharing, sensitive, confidential, details, public | 8.33% |
| 7 | **AI security and risk guidance** | data, security, tools, artificial, intelligence, level, gen, recommendations, information, guidance | 8.33% |
| 8 | **AI agents and automation** | agents, tools, coding, review, automation, systems, ai, development, automated, approval | 8.33% |
| 9 | **Campus tool agreements** | link, external, use, unit, campus, agreements, resources, tools, through, click | 8.33% |
| 10 | **Prompting and limitations** | prompt, example, remember, ask, caution, limitation, language, directly, use, want | 8.33% |

**Table B3:** Topics extracted from the text documents of AI policies at university-level. Topic modeling extracts the word distributions across each topic, and we synthesize the name of each topic using the keywords present in the topic distribution. This particular topic modeling combines the strengths of transformers and c-TF-IDF approached to create dense clusters that allows for identifying interpretable topics while keeping important words in the topic descriptions. We leverage Python-based API to perform the topic extraction automatically.

| ID | Theme | Vocabulary | % of policies |
|---|---|---|---|
| 1 | **Data security and procurement** | data, services, security, level, use, contract, procurement, information, tool, input | 21.95% |
| 2 | **GenAI policy and faculty guidance** | tools, generative, community, information, use, policy, guidance, faculty, external, consider | 12.20% |
| 3 | **Campus tool agreements** | link, external, use, unit, campus, agreements, through, click, contracts, tools | 7.3% |
| 4 | **IP and generative AI output rights** | output, generative, property, intellectual, may, rights, otherwise, party, original, decisions | 7.3% |
| 5 | **Academic integrity** | course, academic, honor, integrity, instructor, instructors, acknowledgement, assignment, students, violation | 7.3% |
| 6 | **Tutoring and academic support** | help, room, tutoring, use, instructors, research, support, chemistry, resources, center | 4.9% |
| 7 | **AI use exceptions and phishing risks** | generative, local, phishing, system, use, policies, students, work, classes, exceptions | 4.9% |
| 8 | **AI content accuracy and publishing** | generated, content, ai, accuracy, libraries, publish, material, example, tools, generative | 4.9% |
| 9 | **AI detection in courses** | students, instructors, tools, courses, course, detection, value, making, learning, however | 4.9% |
| 10 | **Confidential research protection** | unpublished, research, researchers, subjects, work, confidential, may, funding, peer, prohibit | 4.9% |
| 11 | **Enterprise AI and tools** | artificial, intelligence, tools, generative, microsoft, course, copilot, patterns, integrated, powerful | 4.9% |
| 12 | **Institutional AI compliance** | information, institute, including, complies, risk, policies, research, tool | 4.9% |